\documentclass[10pt,twocolumn,letterpaper]{article}

\usepackage{cvpr}
\usepackage{times}
\usepackage{epsfig}
\usepackage{graphicx}
\usepackage{subfigure}
\usepackage{wrapfig}
\usepackage{amsmath}
\usepackage{amssymb}
\usepackage{booktabs}       
\usepackage{stfloats}

\usepackage[pagebackref=true,breaklinks=true,letterpaper=true,colorlinks,bookmarks=false]{hyperref}

\cvprfinalcopy 

\def\cvprPaperID{5187} 

\ifcvprfinal\fi
\begin{document}

\title{PointSIFT: A SIFT-like Network Module for 3D Point Cloud Semantic Segmentation}

\author{Mingyang Jiang\\
Shanghai Jiao Tong University\\
{\tt\small jmydurant@sjtu.edu.cn}
\and
Yiran Wu\\
Shanghai Jiao Tong University\\
{\tt\small yiranwu@sjtu.edu.cn}
\and
Tianqi Zhao\\
Tsinghua University\\
{\tt\small zhaotq16@mails.tsinghua.edu.cn}
\and
Zelin Zhao\\
Shanghai Jiao Tong University\\
{\tt\small sjtuytc@sjtu.edu.cn}
\and
Cewu Lu\\
Shanghai Jiao Tong University\\
{\tt\small lucewu@sjtu.edu.cn}
}


\maketitle

\begin{abstract}
Recently, 3D understanding research sheds light on extracting features from point cloud directly \cite{qi2016pointnet, qi2017pointnetplusplus}, which requires effective shape pattern description of point clouds. Inspired by the outstanding 2D shape descriptor SIFT \cite{sift}, we design a module called PointSIFT that encodes information of different orientations and is adaptive to scale of shape. Specifically, an orientation-encoding unit is designed to describe eight crucial orientations, and  multi-scale representation is achieved by stacking several orientation-encoding units. PointSIFT module can be integrated into various PointNet-based architecture to improve the representation ability. Extensive experiments show our PointSIFT-based framework outperforms state-of-the-art method on standard benchmark datasets. The code and trained model will be published accompanied by this paper. 
\end{abstract}

\section{Introduction}
3D point cloud understanding is a long-standing problem. Typical tasks include 3D object classification \cite{3Dshapenet}, 3D object detection \cite{2Ddriven, qi2017frustum, DeepSlidingShapes} and 3D semantic segmentation \cite{qi2016pointnet, qi2017pointnetplusplus,Riegler2017OctNet}. Among these tasks, 3D semantic segmentation which assigns semantic labels to points is relatively challenging. Firstly, the sparseness of point cloud in 3D space makes most spatial operators inefficient. Moreover, the relationship between points is implicit and difficult to be represented due to the unordered and unstructured property of point cloud. In retrospect of previous work, several lines of solutions have been proposed to resolve the problem. In \cite{supervoxel} handcrafted voxel feature is used to model geometric relationship, and in \cite{scenecut} 2D CNN features from RGBD images are extracted. Additionally, there is a dilemma between 2D convolution and 3D convolution: 2D convolution fails to capture 3D geometry information such as normal and shape while 3D convolution requires heavy computation.

\begin{figure*}[!t]
	\centering
	\includegraphics[width=\linewidth]{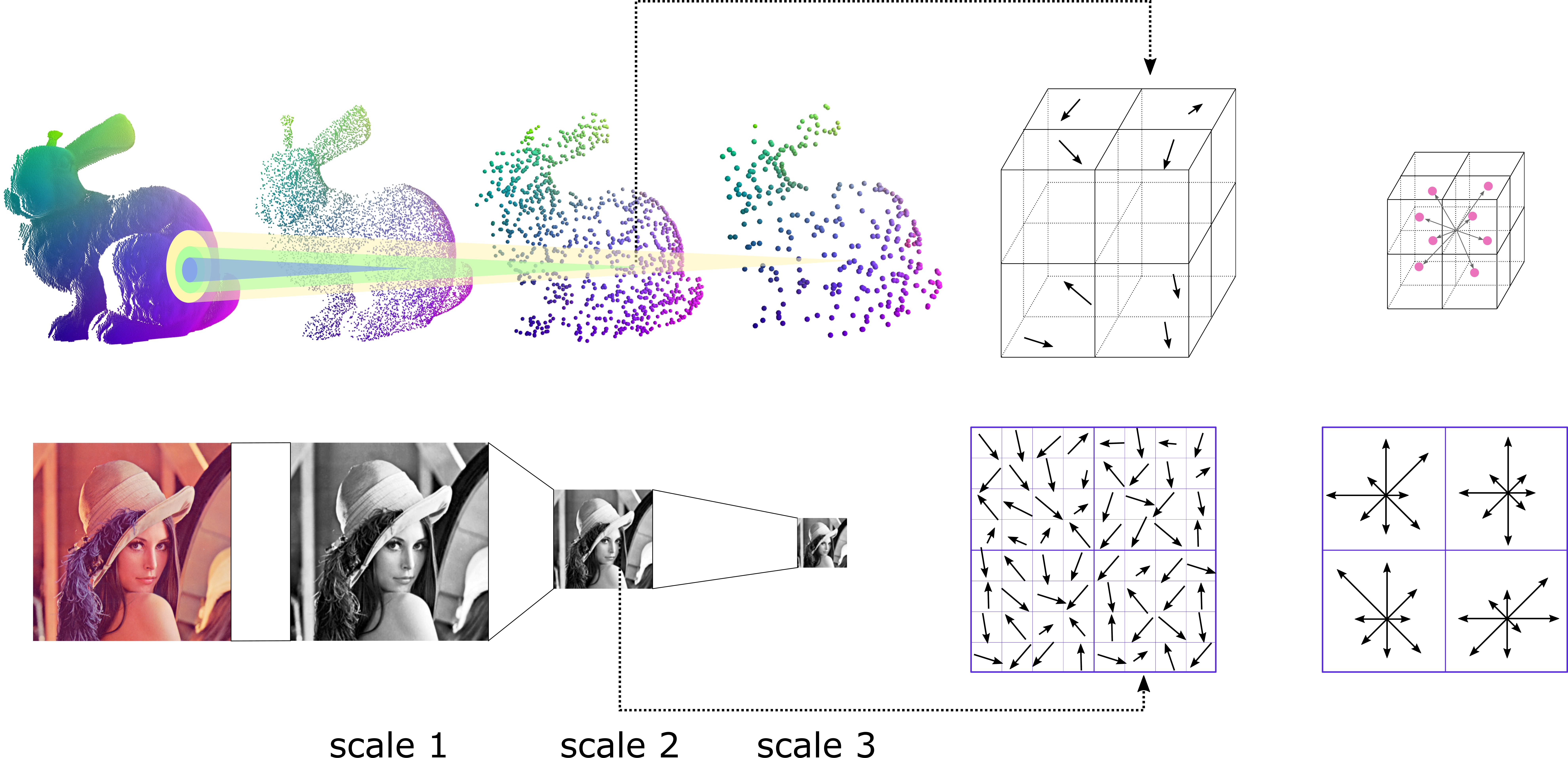}
	\caption{Structure of SIFT \cite{sift} and our PointSIFT module.  The left side shows that both of them can capture multi-scale patterns and is adaptive to various scales. The right side shows that orientation is encoded in each key point/pixel.}  
	\label{fig:sift}
\end{figure*}

Recently, PointNet architecture \cite{qi2016pointnet} directly operates on point cloud instead of 3D voxel grid or mesh. It not only accelerates computation but also notably improves the segmentation performance. In this paper, we also work on raw point clouds. We get inspiration from the successful feature detection algorithm Scale-invariant feature transform (SIFT) \cite{sift} which involves two key properties: scale-awareness and orientation-encoding. Believing that the two properties should also benefit 3D shape description, we design a novel module called PointSIFT for 3D understanding that possesses the properties. The main idea of PointSIFT is illustrated in Figure \ref{fig:sift}. Unlike SIFT algorithm which uses handcrafted features, our PointSIFT is a parametric deep learning module which can be optimized and adapted to point cloud segmentation.

The basic building block of our PointSIFT module is an orientation-encoding (OE) unit which convolves the features of nearest points in 8 orientations. In comparison to K-nearest neighbor search in PointNet++ \cite{qi2017pointnetplusplus} where $K$ neighbors may fall in one orientation, our OE unit captures information of all orientations. We further stack several OE units in one PointSIFT module for representation of different scales. In order to make the whole architecture scale-aware, we connect these OE units by shortcuts and jointly optimize for adaptive scales. Our PointSIFT module receives point cloud with $n$ features each point and outputs points of $n$ features with better representation power. PointSIFT is a general module that can be integrated into various PointNet-based architectures to improve 3D shape representation. 

We further build a hierarchical architecture that recursively applies the PointSIFT module as local feature descriptor. Resembling conventional segmentation framework in 2D \cite{fcn} and 3D \cite{qi2017pointnetplusplus}, our two-stage network first downsamples the point cloud for effective calculation and then upsamples to get dense predictions. The PointSIFT module is used in each layer of the whole framework and significantly improves the representation ability of the network.

Experimental results show that our architecture based on PointSIFT module outperforms state-of-the-art methods on S3DIS\cite{s3Dis} (relative $\mathbf{12\%}$ improvement) and ScanNet\cite{scannet} dataset(relative $\mathbf{8.4\%}$ mean IoU improvement).

\section{Related Work}
Deep learning on 3D data is a growing field of research. We investigate segmentation methods on several important 3D representations. Point cloud segmentation is discussed in more detail. After that, we briefly survey the SIFT descriptor where we borrow inspiration.
\subsection{3D Representation}
\paragraph{Volumetric Representation} The first attempt to apply deep learning is through volumetric representation. Many works \cite{voxelnet,qi2016volumetric,3Dshapenet} attempt to voxelize 3D point cloud or scene into regular voxel grids. However, the main challenges in volumetric representation are data sparsity and computational overhead of 3D convolution. A practical choice for resolution of voxel grid can be $32 \times 32 \times 32$, which is far from sufficient to faithfully represent complex shapes or scenes. Further, conversion is required to construct volumetric grids based on handy data format such as point cloud, which suffers from both truncation error and information loss. While some recent work \cite{fpnn, Riegler2017OctNet, octtree_gen} propose techniques to address the sparsity issue (\eg, octree data structure), the performance of volumetric methods is still not on par with methods based on point cloud \cite{qi2016volumetric}.

\paragraph{Polygonal Meshes} Some literature \cite{spectralnetwork,Geodesic, syncspeccnn} focuses on the use of graph Laplacian to process meshes. Further, functional map and cycle consistency \cite{peter_functional_map, peter_cycle_consistency} helps build correspondence between shapes. However, this kind of methods are confined to manifold meshes.

\paragraph{Multi-view Representation} \cite{qi2016volumetric, multi_view, 3D_assistied} make an effort to exploit the strong capacity of 2D CNNs in 3D recognition. In these works, in order for the input to fit in 2D CNNs, the projection of 3D shapes to 2D images is required. The 3D-level understanding of object (or scene) is achieved by combining 2D images taken from various viewpoints. However, such kind of projection results in the loss of most crucial and discriminative geometric details. For example, calculating normal vectors becomes nontrivial, spatial distance is not preserved, and occlusion prevents a holistic understanding of both local and global structure. Failure to include those geometric details could substantially limit the performance in tasks such as shape completion and semantic segmentation.

\subsection{Deep Learning on Point Cloud}

\paragraph{PointNet and follow-up works} Recently, a series of works propose several effective architectures that process point cloud directly. Among those, a big branch of works apply PoinetNet \cite{qi2016pointnet} as an unordered global or local descriptor. PointNet \cite{qi2016pointnet} is a pioneering effort that applies deep learning to unordered point clouds by point-wise encoding and aggregation through global max pooling. PointNet++ \cite{qi2017pointnetplusplus} proposes a hierarchical neural network to capture local geometric details. PointCNN \cite{pointcnn} uses $\mathcal{X}$-Conv layer instead of vanilla mlp (multilayer perceptron) layer to exploit certain canonical ordering of points. Dynamic Graph CNN \cite{dgcnn} (DGCNN) suggests an alternative grouping method to ball query used in PointNet++ \cite{qi2017pointnetplusplus}: KNN w.r.t. Euclidean distance between feature vectors. Superpoint Graphs\cite{SpGraph} (SPG) first partitions point cloud into superpoints and embed every superpoint with shared PointNet. The semantic labels of superpoints are predicted from the PointNet embedding of current superpoint and spatially neighboring superpoints.  While achieving leading results, we feel that segmentation algorithms could benefit from having ordered operators to some extent.

\paragraph{Rotation Equivariance and Invariance} Another branch of work focuses on rotation equivariance or invariance. G-CNN \cite{gcnn} designs filter so that the filter set is closed under certain rotations (\eg, 90-degree rotation) to achieve rotation invariance for those fixed rotations. This kind of method achieves exact rotation invariance only for some discrete rotations while introduces huge computational overhead. The time complexity is proportional to the cardinality of equivalence classes under the rotations one want to be equivariant of, making it impractical to consider rotation equivariance for large and general groups. Spherical CNN \cite{SphericalCNN_uva, SphericalCNN_upenn} projects 3D shapes onto spheres and process the signal with spherical filters for rotation equivariant representation. Global max pooling that transforms sphere to a single value further achieves rotation invariance. While spherical CNNs is fully invariant to rotation, the projection of shapes onto sphere introduces large error and is not appropriate for objects with certain topological property or scenes. Besides, the operation that helps achieve rotation invariance substantially limits the model capacity, discouraging its use on segmentation tasks.

\subsection{Scale-Invariant Feature Transform (SIFT)}
SIFT \cite{sift} is a local image pattern descriptor widely used in object recognition, 3D modeling, robotics and various other fields. SIFT and its variants \cite{surf} consist of two entities, a scale-invariant detector and a rotation-invariant descriptor.

We borrow inspiration from both entities in SIFT algorithm. In keypoint detection stage, the SIFT algorithm achieves scale invariance with multi-scale representation which we also use for robustly processing objects of various scale. As for feature description stage, SIFT detects dominant orientations for rotation invariance and comprehensively perceives image pattern in different orientations. Given the fact that ordered descriptors like the descriptor of SIFT or kernels of CNN yield impressive results for 2D images, we expect that having such descriptors may also benefit representation of point cloud. The above two observations from SIFT algorithm lead us to design a scale-aware descriptor that encodes information from different orientations with ordered operations.

\section{Problem Statement}
    We first formulate the task of point cloud semantic segmentation. The given point cloud is denoted as $\mathbf{P}$ which is a point set containing $n$ points $p_{1}, p_{2}, ..., p_{n} \in \mathbb{R}^{d}$ with $d$ dimensional feature. The feature vector of each point $p_i$ can be its coordinate $(x_i, y_i, z_i)$ in 3D space (or plus optional feature channels such as RGB values, normal, a representation vector in intermediate step, etc) . 
    The set of semantic labels is denoted as $\mathbf{L}$. A semantic segmentation of a point cloud is a function $\Psi$ which assigns semantic labels to each point in the point cloud. i.e.,
    \begin{equation}
    \Psi: \mathbf{P} \longmapsto L^n
    \end{equation}
    The objective of segmentation algorithms are finding optimal function that gives accurate semantic labels.
    
    Several properties of point set $P$ have been emphasized in previous work \cite{qi2016pointnet,qi2017pointnetplusplus}. The density of $P$ may not be uniform everywhere, and $P$ can be very sparse. Moreover, $P$ as a set is unordered and unstructured which distinguishes point cloud with common sequential or structured data like image or video.

\section{Our Method}
\label{sec:main}
\begin{figure*}[!h]
	\centering
	\includegraphics[width=\linewidth]{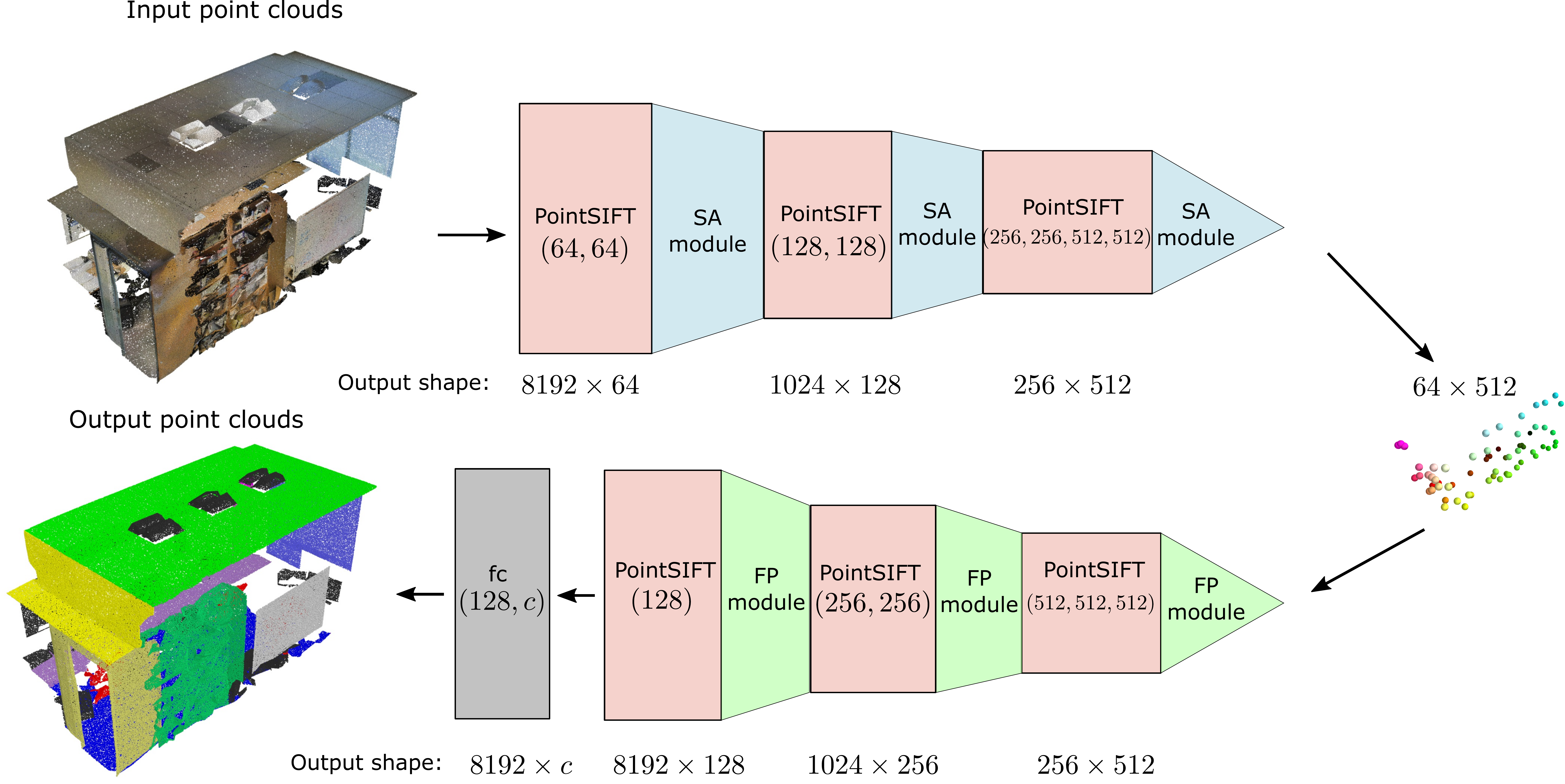}
	\caption{Illustration of our two-stage network architecture. The network consists of downsampling (set abstraction) and upsampling (feature propagation) procedures. PointSIFT modules (marked in red) are interleaved with downsampling (marked in blue) and upsampling (marked in green) layers. Both SA and FP module are introduced in \cite{qi2017pointnetplusplus}. The FP-shortcuts are not shown in the figure for better clarity. PointSIFT$(\cdot)$ specifies feature dimensionalities of each orientation-encoding(OE) units, for example, PointSIFT$(64, 64)$ stands for two stacked OE units both having 64 output feature channels. The number beneath layers is the shape of output point set of corresponding layers, for example, $8192 \times 64$ means 8192 points with 64 feature channel each point.} 
	\label{fig:whole_network}
\end{figure*}

Our network follows a encode-decode (downsample-upsample) framework similar to general semantic segmentation network \cite{segnet} for point cloud segmentation  (Illustrated in Figure \ref{fig:whole_network}) . In the downsampling stage, we recursively apply our proposed PointSIFT module combined with set abstraction (SA) module introduced in  \cite{qi2017pointnetplusplus} for hierarchical feature embedding. For upsampling stage dense feature is enabled by effectively interleaving feature propagation (FP) module \cite{qi2017pointnetplusplus} with PointSIFT module. One of our main contribution and core component of our segmentation network is the PointSIFT module which is endowed with the desired property of orientation-encoding and scale-awareness.

\subsection{PointSIFT Module}
\label{sec:SITT}
Given an $n \times d$ matrix as input which describes a point set of size $n$ with $d$ dimension feature for every point, PointSIFT module outputs an $n \times d$ matrix that assigns a new $d$ dimension feature to every point. 

Inspired by the widely used SIFT descriptor, we seek to design our PointSIFT module as a local feature description method that models various orientations and is invariant to scale.

\subsubsection{Orientation-encoding}
\label{sec:OEC}
Local descriptors in previous methods typically apply unordered operation (\eg, max pooling \cite{qi2017pointnetplusplus, dgcnn}) based on the observation that point cloud is unordered and unstructured. However, using ordered operator could be much more informative (max pooling discards all inputs except for the maximum) while still preserves the invariance to order of input points. One natural ordering for point cloud is the one induced by the ordering of the three coordinates. This observation leads us to the Orientation-encoding(OE) unit which is a point-wise local feature descriptor that encodes information of eight orientations.

\begin{figure}[htbp]
	\centering
	\includegraphics[width=\linewidth]{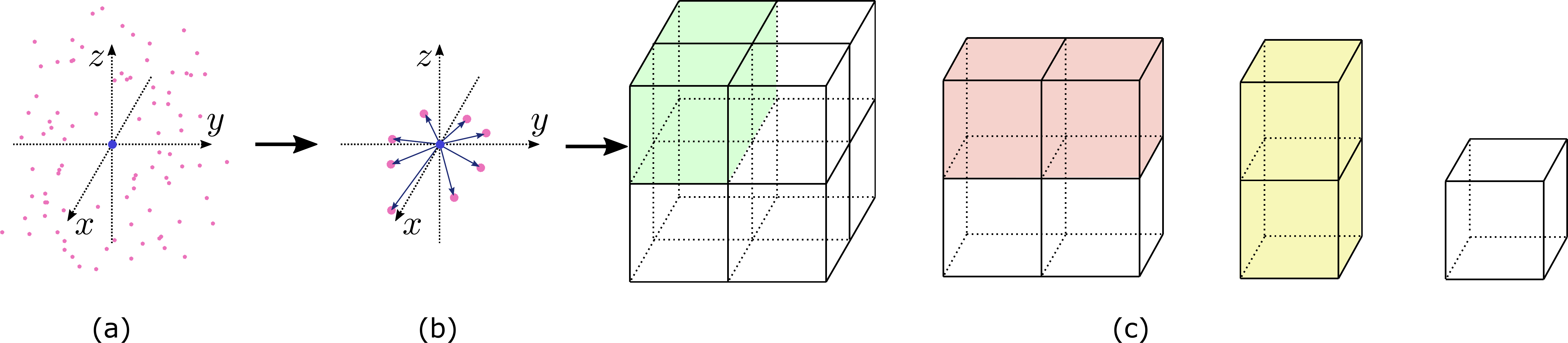}
	\caption{Illustration of Orientation-encoding(OE) Unit. (a): Point cloud in 3D space, the input point is at origin. (b) nearest neighbor search in eight octants. (c) convolution along $X, Y, Z$ axis.} 
	\label{fig:ori_conv}
\end{figure}

 The input of OE unit is a $d$-dimension feature vector $f_0 \in \mathbb{R} ^{d}$ of point $p_0$. Information from eight orientations are integrated by a two-stage scheme to produce an orientation-aware feature $f_0'$. The OE Unit is illustrated in Figure \ref{fig:ori_conv}.

The first stage of OE embedding is Stacked 8-neighborhood(S8N) Search which finds nearest neighbors in each of the eight octants partitioned by ordering of three coordinates. Since distant points provides little information for description of local patterns, when no point exists within searching radius $r$ in some octant, we duplicate $p_0$ as the nearest neighbor of itself.

We further process features of those neighbors which resides in a $2 \times 2 \times 2$ cube for local pattern description centering at $p_0$. Many previous works ignore the structure of data and do max pooling on feature vectors along $d$ dimensions to get new features. However, we believe that ordered operators such as convolution can better exploit the structure of data. Thus we propose orientation-encoding convolution which is a three-stage operator that convolves the $2 \times 2 \times 2$ cube along $X$, $Y$, and $Z$ axis successively. Formally, the features of neighboring points is a vector $V$ of shape $2 \times 2 \times 2 \times d$, where the first three dimensions correspond to three axes. Slices of vector $M$ are feature vectors, for example $M_{1,1,1}$ represents the feature from top-front-right octant.The three-stage convolution is formulated as:
\begin{equation*}
\label{eq:conv}
\begin{aligned}
V_x &= g(Conv(W_x,V)) \in \mathbb{R}_{1 \times 2 \times 2 \times d} \\
V_{xy} &= g(Conv(W_y,V_x)) \in \mathbb{R}_{1 \times 1 \times 2 \times d} \\
V_{xyz} &= g(Conv(W_z,V_{xy})) \in \mathbb{R}_{1 \times 1 \times 1 \times d}\\
\end{aligned}
\end{equation*}
where $W_x\in \mathbb{R}_{2 \times 1 \times 1 \times d}$, $W_y \in \mathbb{R}_{1 \times 2 \times 1 \times d}$ and $W_z\in \mathbb{R}_{1 \times 1 \times 2 \times d}$ are weights of convolution operator (bias is omitted for clarity). In this paper, we set $g(\cdot) = ReLU(\cdot)$. Finally, OE convolution outputs a $d$ dimension feature by reshaping $V_{xyz} \in \mathbb{R}_{1 \times 1 \times 1 \times d}$. Orientation-encoding Convolution (OEC) integrates information from eight spatial orientations and obtains a representation that encodes orientation information.

\subsubsection{Scale-awareness}

\begin{figure}[t]
	\centering
	\includegraphics[width=0.9\linewidth]{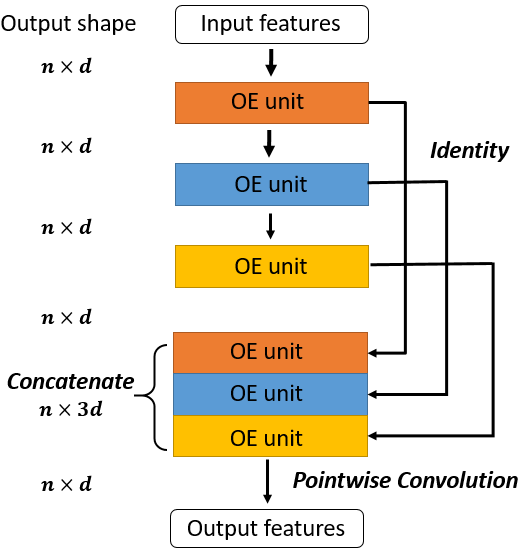}
	\caption{PointSIFT Module. Input features first pass through a series of Orientation-encoding(OE) layers, then outputs of OE units are concatenated and transformed by another point-wise convolution to obtain multi-scale feature.}
	\label{fig:sift_module}
\end{figure}

In order for our PointSIFT module to be scale-aware, we follow the long-standing method of multi-scale representation by stacking several Orientation-encoding (OE) units in PointSIFT module, as Figure \ref{fig:sift_module} illustrated. Higher level OE units have larger receptive field than those of lower level. By constructing a hierarchy of OE units, we obtain a multi-scale representation of local regions in the point cloud. The features of various scales are then concatenated by several identity shortcuts and transformed by another point-wise convolution that outputs a $d$ dimensional multi-scale feature. In the process of jointly optimizing feature extraction and the point-wise convolution that integrates multi-scale feature, neural networks will learn to select or attend to appropriate scales which makes our network scale-aware.

The fact that the input and output vector of our PointSIFT module is of the same shape makes it convenient for our module to be integrated into other existing point cloud segmentation architectures. We are looking forward to future applications of PointSIFT module, probably not restricted to point cloud segmentation domain.

\subsection{Overall Architecture}
\label{sec:all}
We first revisit set abstraction (SA) and feature propagation (FP) module in PointNet++, thus present how to construct our model by SA, FP and pointSIFT modules. 

\subsubsection{Revisit SA and FP Module in PointNet++}

Set abstraction (SA) and feature propagation (FP) modules are proposed in PointNet++ \cite{qi2017pointnetplusplus} that correspond to downsampling and upsampling of point cloud respectively. We give a very brief introduction of SA and FP module here.

A set abstraction module takes in $N \times d$ input standing for a point cloud of $N$ points and $d$ dimensional feature each point. The output is $N' \times d'$ which corresponds to $N'$ downsampled points each with $d'$ dimensional feature. The downsampling is implemented by finding $N'$ centroids with farthest point sampling, assigning points to centroids and then calculate embedding of centroids by feeding feature of assigned points through shared PointNet.

The feature propagation module uses linear interpolation weighted by distances to upsample the point cloud. It receives input point set of size $N$ and outputs an upsampled set of size $N'$ where feature dimensionality is kept same. The upsampling process takes points that are dropped during downsampling, and assign features to them based on features of $k$ nearest points that are not dropped weighted by Euclidean distance in 3D space.

\subsubsection{Architecture Details}

 The input of our architecture is the 3D coordinates (or concatenating with RGB value) of $8192$ points.  In the downsampling stage, following \cite{qi2017pointnetplusplus}, multi-layer perceptron (MLP) is used to transform the input 3D (or 6D) vectors into features with $64$ dimensions. Three consecutive downsampling (set abstraction, SA) operations shrink the size of the point set to $1024$, $256$, $64$ respectively. For the upsampling part, we use feature propagation(FP) module as proposed by \cite{qi2017pointnetplusplus} for dense feature and prediction. The point set is lifted to $256$, $1024$, $8192$ points respectively by three FP layers which are aligned to its counterpart in the downsampling stage. Our PointSIFT module is inserted between all adjacent SA and FP layers. Finally, point features of the last upsampling layer pass through a fully connected layer for semantic label prediction.

Moreover, we insert FP modules not only between downsampling layers but also from downsampling layers to its counterpart in the encoding stage. We call these links FP-shortcuts for they resemble shortcuts by linking corresponding downsampling and upsampling layers and complements low-level information that might be lost during downsampling. The FP-shortcuts lead to much faster convergence which is proved by many prior works \cite{pointcnn, qi2017pointnetplusplus} that use such shortcut flavor connections . The prediction accuracy is also improved by a considerable margin which is also reported in many works, \eg, residual networks \cite{resnet}.

\begin{figure*}[htbp]
	\centering
	\includegraphics[width=\linewidth]{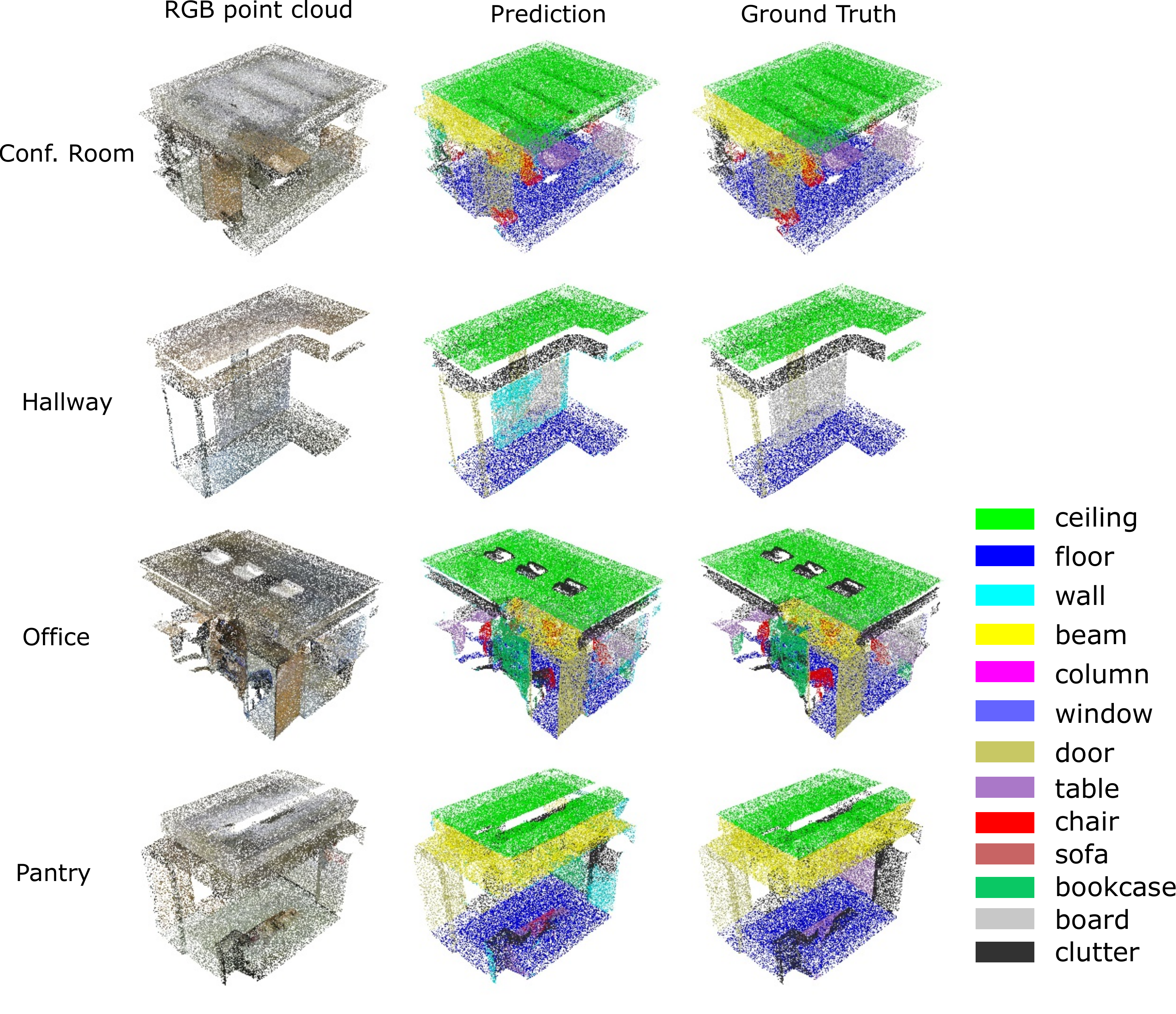}
	\caption{Visualization of results on S3DIS dataset\cite{s3Dis}} 
	\label{fig:experiment}
\end{figure*}

\section{Experiments}
\label{experiments}

Our experiment consists of two parts: verifying the effectiveness of the OE unit and PointSIFT module (Section \ref{analysis}) and introducing the results on semantic segmentation benchmark datasets (Section \ref{result}). In what follows, SA and FP are set abstraction and feature propagation module respectively, which are proposed by \cite{qi2017pointnetplusplus}. 

\subsection{Effectiveness of PointSIFT Module}
\label{analysis}
\paragraph{Orientation-encoding Convolution (OEC)}

\begin{figure}[htbp]
	\centering
	\includegraphics[width=\linewidth]{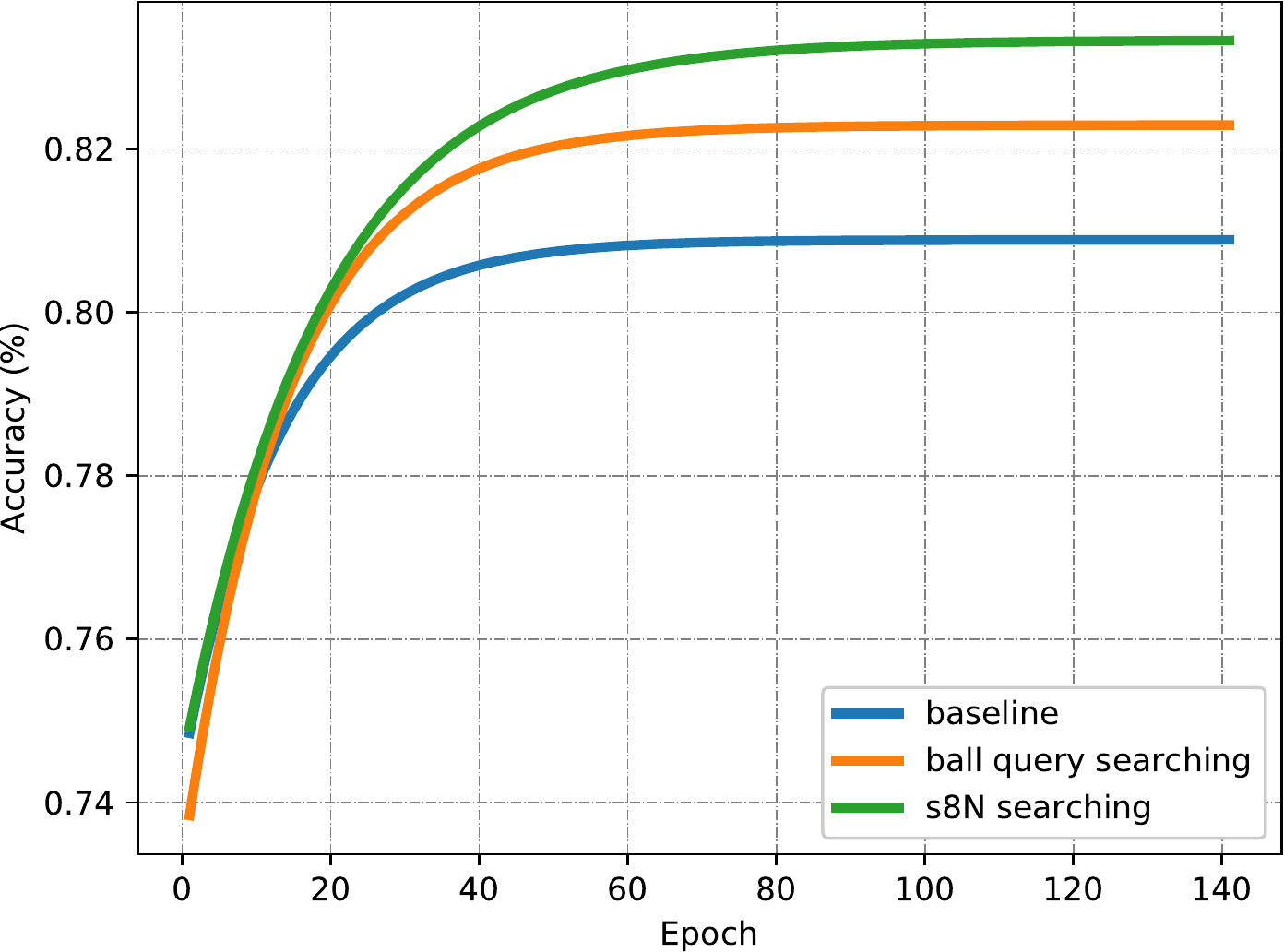}
	\caption{Accuracy for different searching methods. We use Savitzky–Golay filter for smoothing all the lines.}
	\label{fig:cp_net}
\end{figure}
\begin{figure}[!t]
	\centering
	\includegraphics[width=0.45\linewidth]{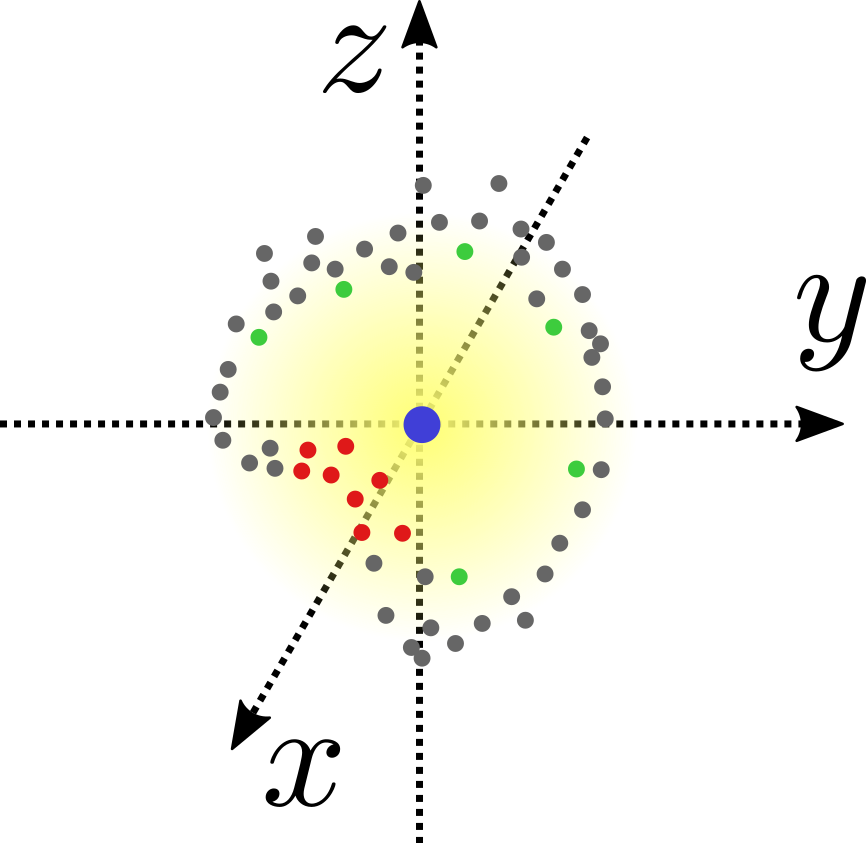}
	\caption{In this case, using K nearest neighbors, all chosen points are from one direction (red points). If we select points in different directions (green points), the representation ability will be better.}
	\label{fig:weakness}
\end{figure}

We apply stacked 8-neighborhood search in OE unit, which is fundamentally different from ball query search proposed in PointNet++ \cite{qi2017pointnetplusplus}. The main difference lies in the fact that S8N search finds neighbors in each of 8 octants, while ball query searches for global nearest neighbors. As Figure \ref{fig:weakness} suggests, the global nearest neighbor search can result in a neighbor set of homogeneous points which is less informative than searching in 8 directions.

To justify our S8N search, we substitute ball query with S8N search in a lightweight version of PointNet++ \cite{qi2017pointnetplusplus} and compare the performance. The detailed architecture of the network is elaborated in Table \ref{table:cp_net}. For fairness, we set the number of neighbors found by the two neighbor search method to be the same. The results are shown in Figure \ref{fig:cp_net} which demonstrates the effectiveness of our S8N grouping plus OE Convolution.

\begin{table*}[htbp]
\setlength{\abovecaptionskip}{10pt}%
	\caption{Effectiveness of PointSIFT Module.}
    \centering
	\begin{tabular}{lcccc}
		\toprule 
		downsampling step & first & second & third & fourth \\
		\toprule
		point cloud size & 8192 & 1024 & 256 & 64 \\
		\midrule
		captured point cloud size of Pointnet++\cite{qi2017pointnetplusplus} & 6570 & 1010 & 255 & 64 \\
		\midrule
		captured point cloud size of PointSIFT framework & 8192 & 1024 & 256 & 64 \\
		\bottomrule
	\end{tabular}
	\label{table:bq}
\end{table*}

Another observation that justifies the PointSIFT module is that more points from individual input point cloud contribute to the final representation for PointSIFT. In PointNet++ \cite{qi2017pointnetplusplus}, the grouping layer fails to assign some points in the point cloud to any centroid and thus lost the information from the unassigned points. We claim the PointSIFT Module can almost avoid information loss in the downsampling process, which is beneficial to semantic segmentation. To prove this, we conduct an experiment on ScanNet \cite{scannet} dataset compared to PointNet++ \cite{qi2017pointnetplusplus}. Given an input of size 8192, the first step is downsampling 8192 points to 1024 points. The method of \cite{qi2017pointnetplusplus} selects 1024 centroids and groups 32 nearest points inside the searching radius.

Our results show that in PointNet++\cite{qi2017pointnetplusplus}, 1622 points on average are not grouped in 8192 points. That is, information of about $20\%$ points are not integrated into the downsampled representation. On the contrary, our PointSIFT module involves more points in computation by performing multiple orientation-encoding(OE) convolutions in OE units. By inserting PointSIFT module before each downsampling layer, our results show that all points can be processed and make a contribution to final predictions. The results are reported in Table \ref{table:bq}.

\begin{table*}[htbp]
\renewcommand\arraystretch{1.2}
\setlength{\abovecaptionskip}{10pt}%
\caption{IoU for all categories of S3DIS\cite{s3Dis} dataset.}
\resizebox{\linewidth}{!}{

	\centering
	\begin{tabular}{l|ccccccccccccc}
	\hline
		Method & ceiling & floor & wall & beam & column & window & door & chair & table & bookcase & sofa & board & clutter \\
	\hline
		PointNet\cite{qi2016pointnet} & 88.0 & 88.7 & 69.3 & 42.4 & 23.1 & 47.5 & 51.6 & 42.0 & 54.1 & 38.2 & 9.6 & 29.4 & 35.2 \\
		SegCloud\cite{SEGCloud} & 90.06 & 96.05 & 69.86 & 0.00 & 18.37 & 38.35 & 23.12 & \textbf{75.89} & 70.40 & 58.42 & 40.88 & 12.96 & 41.60 \\
		SPGraph\cite{SpGraph} & 89.9 & 95.1 & 76.4 & \textbf{62.8} & \textbf{47.1} & 55.3 & 68.4 & 73.5 & 69.2 & \textbf{63.2} & 45.9 & 8.7 & 52.9 \\
	\hline
		Ours & \textbf{93.7} & \textbf{97.9} & \textbf{87.5} & 59.3 & 31.0 & \textbf{73.7} & \textbf{80.7} & 75.1 & \textbf{78.7} & 40.8 & \textbf{66.3} & \textbf{72.2} & \textbf{65.1} \\
	\hline
	\end{tabular}}
	\label{table:id3D_cat}
\end{table*}
\subsubsection{Effectiveness of Scale Awareness}
We design a toy experiment to verify the effectiveness of scale-awareness in our framework. The experiment setting is that we generate 10000 simple shapes (\eg, spheres, cuboids) with different scales, train our framework on generated data. Then we test if the activation magnitude of PointSIFT modules in different layer for certain shape is aligned to the scale of the shape. As aforementioned, different layers in PointSIFT module correspond to different scales. So if such alignment exists we can conclude that the network is aware of scale.
It turns out that 89\% of the time, the position of PointSIFT module with the highest activation in the hierarchy is aligned with the relative scale of input shape w.r.t max and min scale. This toy experiment demonstrates that the proposed PointSIFT framework is aware of scale in some sense. 
 
\newcommand{\tabincell}[2]{$\left\{ \begin{tabular}{#1}#2 \end{tabular}\right\}$}

\begin{table*}[htbp]
\setlength{\belowcaptionskip}{10pt}%
	\centering
	\begin{tabular}{ccccc}
	\hline
		layer name & output size & baseline model & ball query sampling & PointSIFT sampling \\
	\hline
		conv\_1 & 1024$\times$128 & \multicolumn{3}{c}{SA module} \\
	\hline
		conv\_2 & 256$\times$256 & SA module &\tabincell{c}{ball query sampling \\ point-wise convolution\\ SA module} & \tabincell{c}{PointSIFT module \\ (128, 128) \\ SA module} \\
	\hline
		conv\_3 & 64$\times$512 & SA module & \tabincell{c}{ball query sampling \\ point-wise convolution\\ SA module} & \tabincell{c}{PointSIFT module \\ (256, 256) \\ SA module} \\
	\hline
		pf\_conv\_3 & 256$\times$512 & FP module & \tabincell{c}{FP module \\ ball query sampling \\ point-wise convolution } & \tabincell{c}{FP module \\ PointSIFT module \\ (512, 512)} \\
	\hline
		pf\_conv\_2 & 1024$\times$256 & FP module  & \tabincell{c}{FP module \\ ball query sampling \\ point-wise convolution } & \tabincell{c}{FP module \\ PointSIFT module \\ (256, 256)} \\
	\hline
		pf\_conv\_1 & 8192$\times$128 & \multicolumn{3}{c}{FP module} \\
	\hline
		fc & 8192$\times$21 & \multicolumn{3}{c}{fully connected layer} \\
	\hline
	\end{tabular}
	\caption{Architectures for comparison of different sampling methods. After ball query sampling, point-wise convolution takes $32\times1$ kernels for extracting features.}
	
	\label{table:cp_net}
\end{table*}

\subsection{Results on Semantic Segmentation Benchmark Datasets}
\label{result}

\begin{table}[t]
	\renewcommand\arraystretch{1.2}
    \caption{ScanNet\cite{scannet} label accuracy and mIoU}
	\centering
	\begin{tabular}{l|c|c}
	\hline
		Method & Accuracy \% & mean IoU \\
	\hline
		3DCNN\cite{scannet} & 73.0 & -\\

		PointNet\cite{qi2016pointnet} & 73.9 & - \\
	
		PointNet++\cite{qi2017pointnetplusplus} & 84.5 & 38.28 \\
			PointCNN\cite{pointcnn} & 85.1 & - \\
			\hline
			Ours      & \textbf{86.2}  & \textbf{41.5} \\
	        \hline
	\end{tabular}
	\label{table:scannet}
\end{table}

\begin{table}[t]
\renewcommand\arraystretch{1.2}
\setlength{\abovecaptionskip}{10pt}%
\caption{Overall accuracy and meaning intersection over union metric of S3DIS\cite{s3Dis} dataset.}
	\centering
	\begin{tabular}{l|c|c}
	\hline
		Method & Overall Accuracy (\%) & mean IoU (\%) \\
	\hline
		PointNet\cite{qi2016pointnet} & 78.62 & 47.71 \\
		
		SegCloud\cite{SEGCloud} & - & 48.92 \\
		
		SPGraph\cite{SpGraph} & 85.5 & 62.1 \\
		
		PointCNN\cite{pointcnn} & - & 62.74 \\
	
	\hline
		Ours & \textbf{88.72} & \textbf{70.23} \\
	\hline
	\end{tabular}
	
	\label{table:id3D_all}
\end{table}

\paragraph{ScanNet}
ScanNet \cite{scannet} is a scene semantic labeling task with a total of 1513 scanned scenes. We follow \cite{qi2017pointnetplusplus, pointcnn}, use 1201 scenes for training and reserve 312 for testing without RGB information. The result is reported in Table \ref{table:scannet}. Compared with other methods, our proposed PointSIFT method achieves better performance in the sense of per-voxel accuracy. Moreover, our method outperforms PointNet++ even though we do not use multi-scale grouping (MSG) in SA module which helps address the issue of non-uniform sampling density. The result demonstrates the advantage of our PointSIFT module in point searching and grouping.

\paragraph{Stanford Large-Scale 3D Indoor Spaces}
The S3DIS dataset\cite{s3Dis} takes six folders of RGB-D point cloud data from three different buildings (including 271 rooms). Each point is annotated with labels from 13 categories. We prepare the training dataset following \cite{qi2016pointnet} : we split points by room and sample rooms into 1m $\times$ 1m blocks. As have been used in \cite{s3Dis,qi2016pointnet}, we use k-fold strategy for train and test. Overall Accuracy and mIoU are shown in Table \ref{table:id3D_all}. Our PointSIFT architecture outperforms other methods. Per-category IoUs are shown in Table \ref{table:id3D_cat}, our method improves remarkably on results of semantic segmentation task for this dataset and wins in most of the categories. Our method can achieve great results in some hard categories that other methods perform poorly, \textit{i.e.} about 11 mIoU points on the sofa and 42 mIoU points on board. Some results are visualized in Figure \ref{fig:experiment}.


\section{Conclusion}
We propose a novel PointSIFT module and demonstrated significant improvement over state-of-art for semantic segmentation task on standard datasets. The proposed module is endowed with two key properties: First, orientation-encoding units capture information of different orientations. Second, multi-scale representation of PointSIFT modules enables the processing of objects with various scale. Moreover, an effective end-to-end architecture for point cloud semantic segmentation is proposed based on PointSIFT modules. We also conduct comprehensive experiments to justify the effectiveness of the proposed PointSIFT modules.

%
%
%
%
%

{\small
\bibliographystyle{ieee}
\bibliography{cvpr2019}
}

\end{document}